\documentclass[letterpaper, 10 pt, conference]{ieeeconf}  

\IEEEoverridecommandlockouts                              

\usepackage{amsmath} 
\usepackage{amssymb}  
\usepackage{booktabs}
\usepackage{makecell}
\usepackage{multirow}
\usepackage{xcolor}
\usepackage{graphicx}
\usepackage{kotex}
\usepackage{caption}
\usepackage{cleveref}
\usepackage{bbm}

\title{\LARGE \bf RayOcc: Occlusion-Aware Ray Occupancy Estimation via Gaussian Mixture Intensity} 

\author{Junho Kim and Seongwon Lee$^\dagger$
\thanks{J.\,Kim, and S.\,Lee are with the School of Electrical Engineering, Kookmin University, Seoul 02707, South Korea, {\tt\small \{jhk00,sungonce\}@kookmin.ac.kr}.}
\thanks{$^\dagger$ denotes Corresponding Author.}
}

\begin{document}
\newcommand{\red}[1]{{\color{red}#1}}
\newcommand{\todo}[1]{{\color{red}#1}}
\newcommand{\TODO}[1]{\textbf{\color{red}[TODO: #1]}}

\definecolor{nbarrier}{RGB}{255, 120, 50}
\definecolor{nbicycle}{RGB}{255, 192, 203}
\definecolor{nbus}{RGB}{255, 255, 0}
\definecolor{ncar}{RGB}{0, 150, 245}
\definecolor{nconstruct}{RGB}{0, 255, 255}
\definecolor{nmotor}{RGB}{200, 180, 0}
\definecolor{npedestrian}{RGB}{255, 0, 0}
\definecolor{ntraffic}{RGB}{255, 240, 150}
\definecolor{ntrailer}{RGB}{135, 60, 0}
\definecolor{ntruck}{RGB}{160, 32, 240}
\definecolor{ndriveable}{RGB}{255, 0, 255}
\definecolor{nother}{RGB}{139, 137, 137}
\definecolor{nsidewalk}{RGB}{75, 0, 75}
\definecolor{nterrain}{RGB}{150, 240, 80}
\definecolor{nmanmade}{RGB}{213, 213, 213}
\definecolor{nvegetation}{RGB}{0, 175, 0}

\maketitle
\thispagestyle{empty}
\pagestyle{empty}

\begin{abstract}
Camera-only 3D semantic occupancy prediction aims to infer voxel-wise scene semantics from multi-view images, yet remains fundamentally challenging due to depth ambiguity and occlusion. Along a single camera ray, multiple spatially separated surfaces may coexist, making occupancy inherently a multi-label existence problem rather than a single-depth estimation task. However, most existing approaches favor a single dominant depth hypothesis per ray, limiting their ability to model volumetric scenes under complex occlusion. To address this limitation, we introduce RayOcc, an occlusion-aware ray occupancy framework that reformulates ray modeling as multi-label existence prediction. Instead of predicting a categorical depth distribution, RayOcc estimates a non-normalized Gaussian mixture intensity along each ray and converts it into interval-wise occupancy probabilities via a Poisson event formulation, allowing multiple occupied hypotheses to coexist without enforcing mutual competition across depth. The predicted mixture components are interpreted as occupancy hypotheses to initialize sparse 3D Gaussian primitives, which are refined and rasterized for semantic occupancy prediction. Experiments on the nuScenes benchmark show that RayOcc achieves state-of-the-art overall IoU and mIoU among the compared Gaussian-based occupancy methods.
\end{abstract}

\section{INTRODUCTION}

Understanding the full 3D structure of dynamic urban scenes from camera observations is a fundamental problem in autonomous driving~\cite{hu2022uniad, li2022bevdepthacquisitionreliabledepth, chen2017multi, philion2020lss}. Among recent scene representations, 3D semantic occupancy prediction~\cite{cao2022monoscene, jiang2023symphonize, li2023voxformer, li2023fb, selfocc, occnet, wei2023surroundocc, zhang2023occformer, yu2023flashocc} has emerged as a unified and expressive formulation. Instead of detecting objects or estimating a bird’s-eye-view (BEV) map, occupancy prediction aims to recover voxel-wise semantic labels over the entire volumetric space. This representation explicitly models occupied regions, free space, and thin structures, enabling holistic scene reasoning for planning and interaction.

Despite its conceptual elegance, camera-only 3D semantic occupancy prediction remains intrinsically challenging. Unlike LiDAR-based systems~\cite{qi2017pointnetdeephierarchicalfeature, qi2017pointnetdeeplearningpoint, lang2019pointpillarsfastencodersobject, zhou2018voxelnet, liang2022bevfusion, liu2023bevfusion, ye2023lidarmultinet} that directly observe sparse 3D points, vision-based pipelines must infer geometry purely from multi-view images. This inference remains challenging due to depth ambiguity and occlusion. In particular, along a single camera ray, multiple spatially separated surfaces, such as a foreground vehicle, a thin fence behind it, and distant background structures, can project onto the same pixel. Consequently, ray-level occupancy supervision is inherently multi-label, as multiple disjoint depth intervals along a single ray can correspond to occupied regions.

This property distinguishes 3D occupancy prediction from classical depth estimation~\cite{monodepth17, depth_anything_v2, huang2024probabilistic, philion2020lss}. Many depth-based lifting methods assume a single dominant
surface per ray and model depth using a mutually exclusive
categorical distribution. In contrast, volumetric occupancy does not require selecting a single surface location. Instead, it requires determining the independent existence of occupied regions throughout 3D space.

However, most existing camera-based occupancy methods inherit ray modeling strategies originally designed for depth prediction. Specifically, many lifting-based approaches~\cite{philion2020lss, huang2021bevdet, li2022bevdepthacquisitionreliabledepth, Lu_2025_CVPR} discretize depth into bins and predict a softmax-normalized categorical distribution per pixel. Such normalization enforces conservation of probability mass along the ray, implicitly coupling depth intervals through competition. As a result, multiple occupied regions must compete for representational capacity, often leading to suppressed thin structures or fragmented geometry under occlusion.

\begin{figure}
  \centering
  \includegraphics[width=0.48\textwidth]{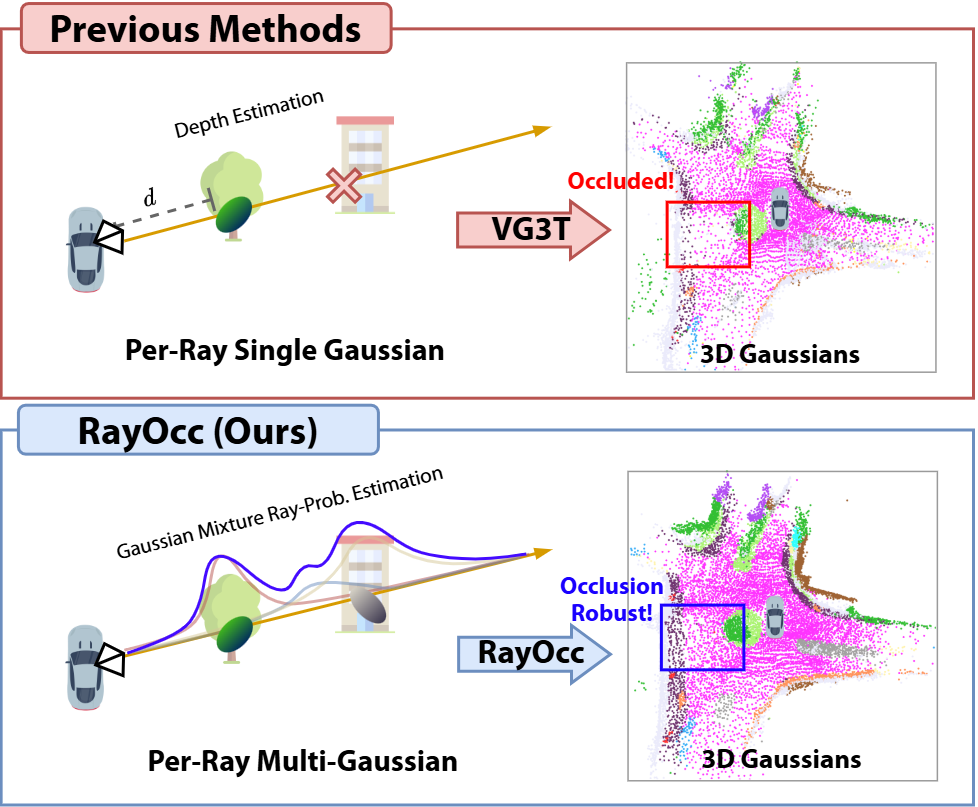}
  \caption{\textbf{Motivation of RayOcc.} RayOcc predicts Mixture Gaussian ray occupancy to initialize multiple Gaussians per ray, improving robustness to occlusion and yielding cleaner Gaussian-to-voxel occupancy predictions than prior single-depth initialization.}
  \vspace{-4mm}
  \label{fig:fig1}
\end{figure}

This mismatch is not merely an implementation artifact but a formulation-level issue. When normalized probabilities are distributed across depth bins, the representation is constrained to favor a dominant hypothesis. Yet volumetric occupancy supervision does not impose such mutual exclusivity, since occupied intervals correspond to independent existence events rather than competing outcomes.

To address this limitation, we propose RayOcc, an occlusion-aware ray occupancy framework that reformulates ray reasoning from depth selection to occupancy existence modeling. Rather than predicting a categorical depth distribution, RayOcc estimates a non-normalized Gaussian mixture intensity along each ray, allowing multiple occupancy hypotheses to coexist without collapsing into a single dominant peak. We interpret occupancy within each depth interval as an event occurrence and convert the integrated mixture intensity into interval-wise existence probabilities via a Poisson event formulation. This design removes the normalization constraint and allows multiple occupied intervals to accumulate evidence independently, better aligning ray modeling with the multi-label nature of volumetric occupancy.

We interpret each mixture component as a candidate occupancy hypothesis along the ray, and convert its amplitude into an activation probability for instantiating a corresponding 3D Gaussian primitive. This adaptive component-wise sampling allows each ray to initialize a variable number of Gaussians, enabling multi-surface representation under depth ambiguity while retaining a sparse primitive-based representation. The initialized primitives are subsequently refined and rasterized to produce voxel-wise semantic occupancy predictions.

Experiments on the nuScenes~\cite{caesar2020nuscenes} benchmark show that RayOcc improves overall occupancy accuracy and Gaussian initialization quality at a moderate additional computational cost.

Our contributions are summarized as follows:
\begin{itemize}
    \item We propose a Mixture Intensity Network that predicts non-normalized Gaussian mixture intensity along each ray and maps integrated intensity to interval-wise existence probabilities for multi-label occupancy modeling.
    \item We introduce adaptive component-wise sampling that converts mixture amplitudes into activation probabilities, enabling a variable number of Gaussian primitives to be initialized per ray.
    \item RayOcc achieves state-of-the-art performance on nuScenes, with improved Gaussian initialization quality at a moderate additional computational cost.
\end{itemize}

\section{Related Work}

\subsection{3D Semantic Occupancy Prediction}

Camera-only 3D semantic occupancy prediction aims to infer dense voxel-wise semantics from multi-view images and has been actively studied as a scalable alternative to LiDAR-based perception. Early camera-based methods rely on dense voxel representations~\cite{li2023voxformer, li2023fb, wei2023surroundocc}, which are expressive but computationally expensive due to the dense modeling of large empty regions. Planar representations such as BEVFormer~\cite{li2022bevformer} and TPVFormer~\cite{huang2023tri} project multi-view image features onto structured 2D grids for more efficient scene understanding. BEV features~\cite{woo2024location} provide an efficient bird's-eye-view representation but lack vertical structure, whereas TPV features partially recover 3D structure through multi-plane representations.

Sparse representations have therefore been explored to reduce spatial redundancy. Point-based methods~\cite{shi2024occupancysetpoints, wang2024opus} provide efficiency but lack spatial extent. Gaussian-based representations~\cite{kong2026rethinking} improve expressiveness by modeling spatially continuous primitives, but their effectiveness depends on accurate Gaussian initialization. GaussianFormer~\cite{huang2024gaussian} initializes Gaussian primitives randomly, whereas GaussianFormer-2~\cite{huang2024probabilistic} and VG3T~\cite{kim2025vg3tvisualgeometrygrounded} initialize a single primitive per pixel, implicitly restricting each ray to a single dominant depth hypothesis. In contrast, our RayOcc models ray-wise occupancy as multi-label existence and initializes multiple Gaussians per ray from mixture-based occupancy hypotheses.

\subsection{Ray-based Density and Occupancy Modeling}

Ray-based lifting is widely used in camera-only perception. Lift-Splat-Shoot-style approaches~\cite{philion2020lss, huang2021bevdet, li2022bevdepthacquisitionreliabledepth, Lu_2025_CVPR} discretize depth into bins and predict a softmax-normalized categorical depth distribution to aggregate image features along camera rays. This formulation is restrictive for semantic occupancy prediction, where multiple depth intervals along a ray may be occupied. The resulting softmax normalization introduces competition across depth bins, which is not well aligned with multi-label volumetric occupancy.

Volume rendering in NeRF~\cite{mildenhall2020nerf, kong2023roomnerf} converts density along a ray into opacity and transmittance for photometric image synthesis. RayOcc uses a related exponential mapping, but for a different purpose, converting integrated mixture intensity into existence probabilities for supervised ray-level occupancy prediction. Unlike rendering weights that are coupled through transmittance, RayOcc predicts interval-wise occupancy probabilities without enforcing visibility-based competition along the ray.

\begin{figure*}
  \centering
  \includegraphics[width=1.0\textwidth]{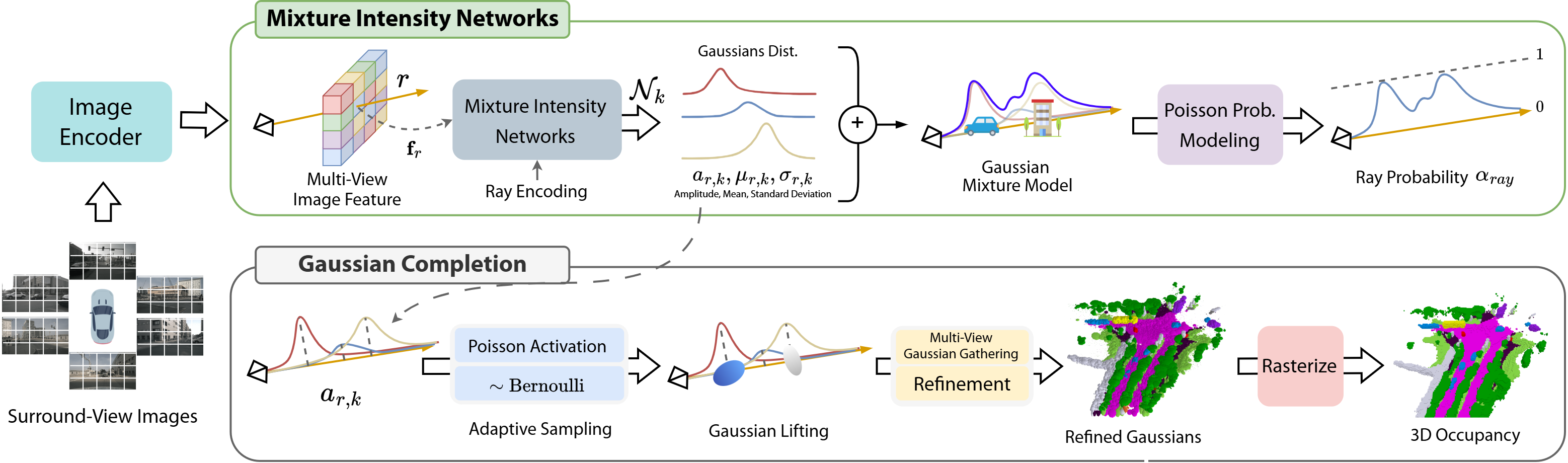}
  \caption{\textbf{Overview of RayOcc.} Multi-view images are encoded into per-ray features and concatenated with ray encoding. A Mixture Intensity Network predicts ray-wise mixture parameters, which are converted into multi-label ray occupancy probabilities via Poisson probability modeling. The predicted mixture weights are used for adaptive sampling to initialize a compact set of semantic 3D Gaussians, which are refined and rasterized to produce the final 3D semantic occupancy grid.}
  \label{fig:fig2}
  \vspace{-5mm}
\end{figure*}

\subsection{Mixture Density Networks}

Mixture Density Networks (MDNs)~\cite{bishop1994mixture} model conditional distributions using mixtures of parametric densities and are commonly used to represent multi-modal continuous predictions. Mixture formulations have been applied in vision tasks such as stereo disparity estimation~\cite{tosi2021smd} and sparse-input novel view synthesis~\cite{Seo_2023_CVPR}.

In contrast, ray-level semantic occupancy is not a continuous regression target but a multi-label existence prediction problem over depth. Along a camera ray, multiple depth intervals may be occupied simultaneously. Standard MDN formulations normalize mixture weights, coupling components through a fixed total probability mass. This coupling is not well aligned with occupancy supervision, where disjoint occupied intervals should not compete. RayOcc instead predicts a non-normalized mixture intensity along each ray and maps it to bin-wise existence probabilities, allowing multiple occupancy hypotheses to coexist.


\section{Proposed Approach}
In this section, we first introduce RayOcc’s Occlusion-Aware Ray Occupancy Estimation via Gaussian Mixture Intensity Network framework. Then, we present the details of the RayOcc model architecture.

\subsection{Problem Setup and Overview}
Vision-centric 3D semantic occupancy prediction aims to recover the geometric structure and semantic labels of a 3D scene from multi-view images. Formally, given a set of multi-view images
$\mathcal{I} = \{ I_i \}_{i=1}^{N}$, 
where $N$ denotes the number of input views. The objective is to predict a voxel-wise semantic occupancy grid $\mathbf{O} \in \mathbb{R}^{C \times X \times Y \times Z}$, where $\{X, Y, Z\}$ represents the spatial resolution of the 3D voxel space and $C$ is the number of predefined semantic classes.

Instead of directly constructing a dense 3D feature volume for all voxels, we adopt a sparse and continuous scene representation based on 3D Gaussian primitives.
Specifically, we represent the scene as a set of $M$ Gaussians $\mathbf{G} = \left\{ g_m \right\}_{m}^{M}, \quad
g_m = \left(\mathbf{p}_m, \mathbf{s}_m, \mathbf{q}_m, \alpha_m, \mathbf{f}_m \right)$, where $\mathbf{p}_m \in \mathbb{R}^3$ denotes the 3D mean of the Gaussian, $\mathbf{s}_m\in\mathbb{R}^3$ as scale, $\mathbf{q}_m\in\mathbb{R}^4$ is a rotation represented as a quaternion, $\alpha_m \in [0,1]$ as an opacity, and $\mathbf{f}_m \in \mathbb{R}^{C}$ a semantic vector. 

\subsection{Ray Occupancy Modeling via Gaussian Mixture Intensity}
Given $N$ multi-view images $\mathcal{I}$, we extract per-view feature maps $\mathbf{F}_n\in\mathbb{R}^{C\times H\times W}$ using a pretrained Visual Geometry Grounded Transformer (VGGT) backbone~\cite{wang2025vggtvisualgeometrygrounded}.
Following VG3T~\cite{kim2025vg3tvisualgeometrygrounded}, we adapt VGGT to the nuScenes by removing camera tokens and retaining only visual tokens for dense prediction.
VGGT refines these tokens through in-frame and cross-frame attention to aggregate contextual information from both within the same view and across views, yielding multi-view-aware tokens.
The refined visual tokens are then projected to spatial feature maps $\mathbf{F}_n$ through a DPT head~\cite{ranftl2021visiontransformersdenseprediction}, which are fed to our Mixture Intensity Networks for ray occupancy estimation.
For each feature map pixel, we form a camera ray $r$ with origin $\mathbf{o}_r\in\mathbb{R}^3$ and direction $\mathbf{u}_r\in\mathbb{R}^3$, and obtain a per-ray feature vector $\mathbf{f}_r\in\mathbf{F}$.
With origin $\mathbf{o}_r$ and direction vector $\mathbf{u}_r$, we encode ray as Pl\"ucker coordinate $\mathbf{e}_r$ and concatenate to the per-ray feature $\mathbf{f}_r$ to inject geometric context:
\begin{equation}
\mathbf{h}_r = [\mathbf{f}_r, \mathbf{e}_r].
\end{equation}

We discretize the depth range $[d_{\min}, d_{\max}]$ into $S$ bins with edges $\{d_i\}_{i=0}^{S}$.
Unlike categorical depth distributions that apply softmax across bins enforcing mutual exclusivity, we model ray occupancy as a {non-normalized event intensity} along depth.
For each ray $r$, a Mixture Intensity Network $f_{\mathrm{mix}}$ predicts the parameters of $K$ Gaussian components $\{a_{r,k}, \mu_{r,k}, \sigma_{r,k}\}_{k=1}^{K}$.
\begin{equation}
a_{r},\mu_{r},\sigma_{r}=f_{\mathrm{mix}}(h_r),
\end{equation}
where $a_{r,k}\ge 0$ is a non-negative amplitude, $\mu_{r,k}\in[d_{\min}, d_{\max}]$ is the component mean, and $\sigma_{r,k}>0$ is the component standard deviation.
We define the continuous ray-wise intensity function:
\begin{equation}
\lambda_r(d) = \sum_{k=1}^{K} a_{r,k}\,\mathcal{N}(d;\mu_{r,k}, \sigma_{r,k}^2).
\end{equation}
Unlike a categorical depth distribution, $\lambda_r$ is not constrained by $\sum_k a_{r,k}=1$, allowing multiple occupancy hypotheses to receive high intensity along a single ray.

We interpret occupancy within each depth bin as an {event occurrence} and convert intensity into the probability that {at least one} occupancy event exists in that bin.
The integrated intensity of bin $i$ is
\begin{equation}
\Lambda_{r,i}
=
\int_{d_{i-1}}^{d_i}\lambda_r(t)\,dt
\approx
\lambda_r(c_i)\Delta_i,
\quad
c_i=\frac{d_{i-1}+d_i}{2}.
\end{equation}
Assuming a Poisson event model, the probability of at least one event in bin $i$ becomes
\begin{equation}
\Pr(N_{r,i}\ge 1)=1-\Pr(N_{r,i}=0),
\end{equation}
\begin{equation}
p_{r,i} = 1 - \exp(-\Lambda_{r,i}).
\end{equation}
The resulting $\alpha_{\mathrm{ray}} = \{p_{r,i}\}_{i=1}^{S}$ forms a multi-label ray occupancy distribution, where multiple depth bins can simultaneously have high probability. This matches semantic occupancy supervision, in which a ray may intersect multiple semantic surfaces.

\begin{table*}[ht]
    \centering
    \renewcommand{\arraystretch}{1}
    \setlength{\tabcolsep}{2.5pt}
    \caption{\textbf{3D semantic occupancy prediction results on nuScenes benchmark.}}
    \begin{tabular}{l|c c | c c c c c c c c c c c c c c c c}
        \toprule
        Method
        & IoU
        & mIoU
        & \rotatebox{90}{\textcolor{nbarrier}{$\blacksquare$} barrier}
        & \rotatebox{90}{\textcolor{nbicycle}{$\blacksquare$} bicycle}
        & \rotatebox{90}{\textcolor{nbus}{$\blacksquare$} bus}
        & \rotatebox{90}{\textcolor{ncar}{$\blacksquare$} car}
        & \rotatebox{90}{\textcolor{nconstruct}{$\blacksquare$} const. veh.}
        & \rotatebox{90}{\textcolor{nmotor}{$\blacksquare$} motorcycle}
        & \rotatebox{90}{\textcolor{npedestrian}{$\blacksquare$} pedestrian}
        & \rotatebox{90}{\textcolor{ntraffic}{$\blacksquare$} traffic cone}
        & \rotatebox{90}{\textcolor{ntrailer}{$\blacksquare$} trailer}
        & \rotatebox{90}{\textcolor{ntruck}{$\blacksquare$} truck}
        & \rotatebox{90}{\textcolor{ndriveable}{$\blacksquare$} drive. suf.}
        & \rotatebox{90}{\textcolor{nother}{$\blacksquare$} other flat}
        & \rotatebox{90}{\textcolor{nsidewalk}{$\blacksquare$} sidewalk}
        & \rotatebox{90}{\textcolor{nterrain}{$\blacksquare$} terrain}
        & \rotatebox{90}{\textcolor{nmanmade}{$\blacksquare$} manmade}
        & \rotatebox{90}{\textcolor{nvegetation}{$\blacksquare$} vegetation}
        \\
        \midrule
        MonoScene~\cite{cao2022monoscene} & 23.96 & 7.31 & 4.03 &	0.35& 8.00& 8.04&	2.90& 0.28& 1.16&	0.67&	4.01& 4.35&	27.72&	5.20& 15.13&	11.29&	9.03&	14.86 \\
        
        Atlas~\cite{murez2020atlas} & 28.66 & 15.00 & 10.64&	5.68&	19.66& 24.94& 8.90&	8.84&	6.47& 3.28&	10.42&	16.21&	34.86&	15.46&	21.89&	20.95&	11.21&	20.54 \\
        
        BEVFormer~\cite{li2022bevformer} & 30.50 & 16.75 & 14.22 &	6.58 & 23.46 & 28.28& 8.66 &10.77& 6.64& 4.05& 11.20&	17.78 & 37.28 & 18.00 & 22.88 & 22.17 & 13.80 & {22.21}\\
        
        TPVFormer~\cite{huang2023tri} & 11.51 & 11.66 & 16.14&	7.17& 22.63	& 17.13 & 8.83 & 11.39 & 10.46 & 8.23&	9.43 & 17.02 & 8.07 & 13.64 & 13.85 & 10.34 & 4.90 & 7.37\\
        
        TPVFormer*~\cite{huang2023tri}  & {30.86} & 17.10 & 15.96&	 5.31& 23.86	& 27.32 & 9.79 & 8.74 & 7.09 & 5.20& 10.97 & 19.22 & {38.87} & {21.25} & {24.26} & {23.15} & 11.73 & 20.81\\
        
        OccFormer~\cite{zhang2023occformer} & {31.39} & {19.03} & {18.65} & {10.41} & {23.92} & \underline{30.29} & {10.31} & {14.19} & {13.59} & {10.13} & {12.49} & {20.77} & {38.78} & 19.79 & 24.19 & 22.21 & {13.48} & {21.35}\\
        
        SurroundOcc~\cite{wei2023surroundocc} & {31.49} & {20.30}  & \textbf{{20.59}} & {11.68} & {28.06} & \textbf{30.86} & {10.70} & {15.14} & \textbf{14.09} & \textbf{12.06} & {14.38} & {22.26} & 37.29 & {23.70} & {24.49} & {22.77} & {14.89} & {21.86}  \\

        GaussianFormer~\cite{huang2024gaussian} & 29.83 & {19.10} & {19.52} & {11.26} & {26.11} & {29.78} & {10.47} & {13.83} & {12.58} & {8.67} & {12.74} & {21.57} & {39.63} & {23.28} & {24.46} & {22.99} & 9.59 & 19.12 \\
        
        GaussianFormer-2~\cite{huang2024probabilistic} & 30.56 & {20.02} & {20.15} & {12.99} & {27.61} & {30.23} & {11.19} & \underline{15.31} & {12.64} & {9.63} & {13.31} & {22.26} & {39.68} & {23.47} & {25.62} & {23.20} & 12.25 & 20.73 \\

        VG3T~\cite{kim2025vg3tvisualgeometrygrounded} & \underline{34.06} & \underline{21.74} & 19.95 & \textbf{13.63} & \underline{28.69} & 29.52 & \underline{12.69} & \textbf{16.02} & \underline{13.77} & \underline{10.64} & \textbf{15.75} & \underline{23.02} & \underline{41.74} & \underline{26.26} & \underline{27.52} & \underline{26.44} & \textbf{16.51} & \textbf{25.75} \\
        
        \midrule
        \textbf{Ours} & \textbf{34.84} & \textbf{22.03} & \underline{20.31} & \underline{13.32} & \textbf{29.98} & 29.79 & \textbf{15.31} & {14.72} & {13.25} & {10.55} & \underline{15.47} & \textbf{23.53} & \textbf{42.67} & \textbf{27.02} & \textbf{28.51} & \textbf{26.96} & \underline{15.69} & \underline{25.44} \\
        
        \bottomrule
    \addlinespace[0.4mm]
    \multicolumn{19}{l}{* means supervised by dense occupancy annotations
 as opposed to the original LiDAR segmentation labels.} \\
    \addlinespace[0.4mm]
    \multicolumn{19}{l}{The best and second-best performances are represented by \textbf{bold} and \underline{underline} respectively.} \\
    \end{tabular}
    \label{table:nuscenes_results}
    \vspace{-3mm}
\end{table*}

\subsection{Adaptive Component-Wise Sampling}
Given the predicted mixture components, we construct an initial set of 3D Gaussian positions by interpreting each mixture component as an independent occupancy hypothesis.
We convert the non-negative amplitude $a_{r,k}$ into a component activation probability
\begin{equation}
q_{r,k} = 1 - \exp(-\beta a_{r,k}),
\label{eq:component_activation}
\end{equation}
where $\beta$ controls the expected number of selected hypotheses per ray.
We then sample a binary gate $z_{r,k} \sim \mathrm{Bernoulli}(q_{r,k})$. This allows each ray to instantiate a variable number of Gaussians, enabling multi-hypothesis depth modeling.
Let $\{k' \in \{1,\dots,K\} \mid z_{r,k}=1\}$ denote the index set of selected components for ray $r$.
For each selected component $k'$, we lift its mean depth to a 3D Gaussian position $\mathbf{p}_{r,k'}$ using known camera ray geometry:
\begin{equation}
    \mathbf{p}_{r,k'} = \mathbf{o}_r + \mu_{r,k'}\,\mathbf{u}_r,
\end{equation}
where $\mathbf{o}_r$ and $\mathbf{u}_r$ are the ray origin and direction, respectively, and $\mu_{r,k'}$ is the mean depth of the selected mixture component.
We then apply a lightweight grid-based subsampling operator to remove redundant Gaussians and build a compact sparse Gaussian latent set $\mathcal{G}$, which serves as input to the subsequent refinement and Gaussian rasterization stage.
 
\subsection{Refinement and Rasterization}
Given the initialized Gaussian latent set $\mathcal{G}=\{(\mathbf{p}_m, \mathbf{f}_m)\}_{m=1}^{M}$, we further refine the Gaussian attributes and aggregate them into a 3D Semantic Occupancy.

We adopt a 3D sparse convolutional refinement network $\Phi(\cdot)$ to incorporate local 3D context among neighboring Gaussians and to predict residual updates to their position and estimate other gaussian attributes.
For each Gaussian $g_m$, we predict a residual mean offset $\Delta\mathbf{p}_m$, scale $\mathbf{s}_m$, quaternion $\mathbf{q}_m$, and opacity $\alpha_m$:
\begin{equation}
\Delta\mathbf{p}_m,\ \mathbf{s}_m,\ \mathbf{q}_m,\ \alpha_m,\ \mathbf{f}_m=\Phi(\mathbf{f}_m),
\end{equation}
and complete set of Gaussian by updating the center as
\begin{equation}
\mathbf{G} = (\mathbf{p}_m + \Delta\mathbf{p}_m,\ \mathbf{s}_m,\ \mathbf{q}_m,\ \alpha_m, \mathbf{f}_m).
\end{equation}

After refinement, we obtain the final Gaussian set $\mathbf{G}$.
The voxel-wise semantic occupancy prediction $\hat{\mathbf{O}}\in\mathbb{R}^{C\times X\times Y\times Z}$ is then inferred via a probabilistic Gaussian splatting operator:
\begin{equation}
\hat{\mathbf{O}} = \mathrm{Prob}({\mathbf{G}}),
\end{equation}
where $\mathrm{Prob}(\cdot)$ follows the probabilistic superposition formulation in GaussianFormer-2~\cite{huang2024probabilistic}.

\subsection{Loss}
We train RayOcc end-to-end with a weighted sum of an occupancy loss and an auxiliary ray supervision loss:
\begin{equation}
    L_{total}=\lambda_{occ}L_{occ}+\lambda_{ray}L_{ray},
\end{equation}
where $\lambda_{occ}$ and $\lambda_{ray}$ are scalar weights.

\textbf{Occupancy Loss.}
Due to the imbalance in occupancy grids, we adopt a class-balanced loss that explicitly emphasizes non-empty voxels while still optimizing region-level overlap. Specifically, we combine a weighted voxel-wise cross-entropy term with the Lovász-Softmax loss:
\begin{equation}
    L_{occ}=L_{ce}+L_{lov}.
\end{equation}

\textbf{Ray Distribution Loss.}
To supervise the predicted ray-wise bin occupancy $\alpha_{ray}$, we derive binary ground-truth labels along each camera ray from the dense occupancy annotations.
Given the camera calibration, each pixel corresponds to a ray with origin $\mathbf{o}_r$ and unit direction $\mathbf{u}_r$.
We uniformly discretize the depth range $[d_{\min},d_{\max}]$ into $S$ bins and sample one 3D reference point per bin at the bin center $d_i$:
\begin{equation}
\mathbf{x}_{r,i}=\mathbf{o}_r + d_i\,\mathbf{u}_r,\quad i=1,\dots,S.
\end{equation}
We then query each reference point $\mathbf{x}_{r,i}$ into the occupancy grid and assign a binary label $y_{r,i}\in\{0,1\}$ indicating whether the corresponding voxel is non-empty.
The resulting $y_{r,i}$ provides multi-label supervision over depth bins, allowing multiple occupied intervals along a single ray.

To address the severe class imbalance between empty and occupied bins, we supervise the predicted per-ray bin occupancy $\alpha_{ray}$ using the ground-truth ray labels $y_{r,i}$ with a combination of binary cross-entropy (BCE) and Dice loss:
\begin{equation}
    L_{ray}=L_{bce}+L_{dice}.
\end{equation}

\section{Experiments}

\subsection{Datasets}
We conduct experiments on the nuScenes~\cite{caesar2020nuscenes} dataset, a large-scale public benchmark for autonomous driving that contains 1000 driving scenes collected in Boston and Singapore. The official split comprises 700/150/150 scenes for training, validation, and testing, respectively. Each scene lasts 20 seconds and includes synchronized multi-view RGB images captured by 6 surrounding cameras, with keyframes annotated at 2\,Hz.

For 3D semantic occupancy prediction, we use the dense 3D semantic occupancy annotations provided by SurroundOcc~\cite{wei2023surroundocc} for both training and evaluation. The voxel grid covers $[-50,50]$ meters along the $x$ and $y$ axes and $[-5,3]$ meters along the $z$ axis, with a resolution of $200 \times 200 \times 16$, corresponding to a voxel size of 0.5\,m. Each voxel is assigned one of 18 labels, including 16 semantic classes, 1 empty class, and 1 unknown class.

\subsection{Evaluation Metrics}
We evaluate 3D semantic occupancy prediction using mean Intersection-over-Union (mIoU) and Intersection-over-Union (IoU):
\begin{equation}
{\rm mIoU}=\frac{1}{|\mathcal{C}'|}\sum_{i\in \mathcal{C}'} \frac{TP_i}{TP_i+FP_i+FN_i},
\end{equation}
\begin{equation}
{\rm IoU}=\frac{TP_{\neq c_0}}{TP_{\neq c_0}+FP_{\neq c_0}+FN_{\neq c_0}},
\end{equation}
where $\mathcal{C}'$ denotes the set of non-empty semantic classes, $c_0$ is the empty class, and $TP$, $FP$, and $FN$ denote the numbers of true positive, false positive, and false negative voxels, respectively.

\begin{figure*}
  \centering
  \includegraphics[width=1.0\textwidth]{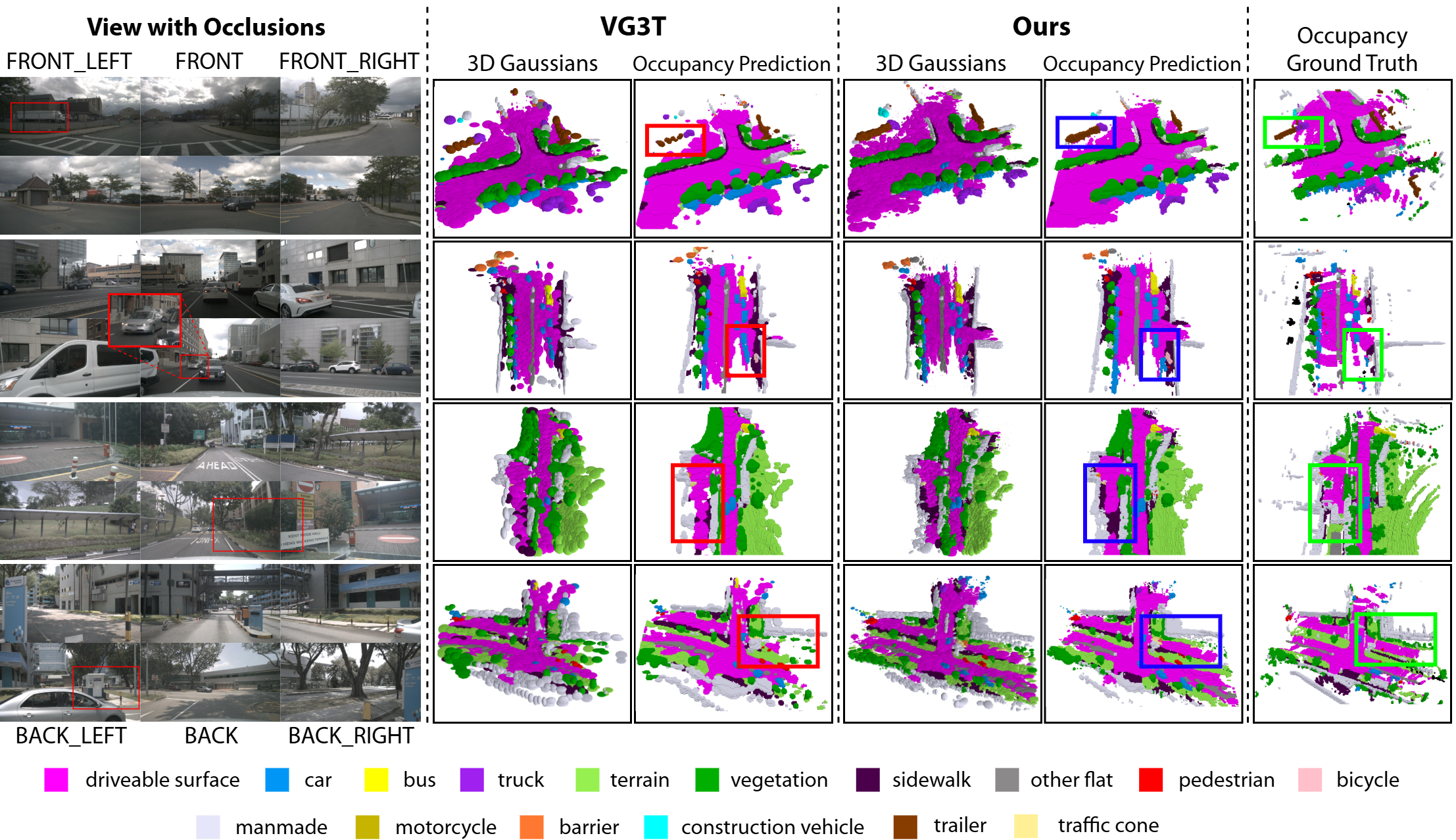}
  \caption{Qualitative results of our method and VG3T on SurroundOcc dataset.}
  \label{fig:fig3}
  \vspace{-3mm}
\end{figure*}

\subsection{Implementation Details}
We set the input image resolution to $900\times1600$ and apply data augmentations, including random horizontal flipping and photometric distortion. 
We employ a pretrained VGGT backbone~\cite{wang2025vggtvisualgeometrygrounded}; each $900\times1600$ image is resized to $294\times518$ before being fed into the backbone to extract image features.
For ray occupancy estimation, we discretize the depth range $[1,71]$ meters into $S=128$ bins and set the number of mixture components in the Mixture Intensity Networks to $K=3$.
For adaptive component-wise sampling, we set the Poisson activation parameter to $\beta=1$ and use stochastic Bernoulli sampling during training, while applying a deterministic thresholding at inference.
We further apply grid-based subsampling with a 0.5\,m cell size to remove redundant Gaussians.
We optimize the model using AdamW~\cite{loshchilov2017adamw} with a weight decay of 0.01.
The learning rate is warmed up in the first 500 iterations to a maximum of $1\times10^{-4}$ and then decayed with a cosine annealing schedule.
We train the model for 20 epochs in bfloat16 precision and use FlashAttention~\cite{dao2023flashattention2fasterattentionbetter} to accelerate attention computation.
All experiments are conducted with a batch size of 4 on NVIDIA L40s GPUs.

\subsection{Main Results}
We compare RayOcc with recent camera-only 3D semantic occupancy methods on the nuScenes~\cite{caesar2020nuscenes} validation set. Table~\ref{table:nuscenes_results} reports the quantitative results. RayOcc achieves the best overall IoU and mIoU among the compared Gaussian-based methods, reaching 34.84 IoU and 22.03 mIoU. Compared with VG3T~\cite{kim2025vg3tvisualgeometrygrounded}, RayOcc improves IoU from 34.06 to 34.84 and mIoU from 21.74 to 22.03. Compared with GaussianFormer-2~\cite{huang2024probabilistic}, RayOcc improves IoU from 30.56 to 34.84 and mIoU from 20.02 to 22.03.
We observe consistent gains on large background categories that dominate scene geometry.
In particular, RayOcc improves over VG3T~\cite{kim2025vg3tvisualgeometrygrounded}, demonstrating stronger modeling of static layout under camera-only geometry ambiguity.
RayOcc also yields competitive performance on foreground objects and improves several challenging categories such as {bus} and {construction vehicle}, while remaining comparable on other dynamic and small classes.
Overall, these results validate that RayOcc’s occlusion-aware ray occupancy modeling provides more reliable geometry prediction by improved IoU.

\begin{table}
    \centering
    \caption{Gaussian Sampling Methods Comparison.}
    \setlength{\tabcolsep}{0.01\linewidth}
    \resizebox{1\linewidth}{!}{
    \begin{tabular}{c|c|cc}
        \toprule
        \multirow{2}{*}{Method} & \multirow{2}{*}{Gaussian Sampling} & \multicolumn{2}{c}{Position}\\
        & & Perc. (\%) \( \uparrow \) & Dist. (m) \( \downarrow \) \\
        \midrule
        GaussianFormer~\cite{huang2024gaussian} & Random & 16.41 & 3.07 \\
        GaussianFormer-2~\cite{huang2024probabilistic} & Categorical & 28.85 & 1.24 \\
        VG3T~\cite{kim2025vg3tvisualgeometrygrounded} & Depth & {51.22} & {0.97} \\
        \midrule
        RayOcc (ours) & Component-wise & \textbf{61.07} & \textbf{0.70} \\
        \bottomrule
    \end{tabular}}
    \label{table:initializing}
    \vspace{-3mm}
\end{table}

\begin{table}
    \centering
    \setlength{\tabcolsep}{0.015\linewidth}
    \scriptsize
    \caption{Efficiency Comparison with Gaussian Represenation Methods.}
    \scalebox{1.05}{
    \begin{tabular}{l|ccc|cc}
        \toprule
        Method & \makecell{Number of\\ Gaussians}$\downarrow$ & \makecell{Latency\\ (ms)}$\downarrow$ & \makecell{Memory\\ (MB)}$\downarrow$ & mIoU & IoU \\
        \midrule
        GaussianFormer~\cite{huang2024gaussian} & 144000 & {372} & 6229 & 19.10 & 29.83 \\ 
        GaussianFormer-2~\cite{huang2024probabilistic} & 25600 & 513 & {3063} & 20.02 & 30.56 \\
        VG3T~\cite{kim2025vg3tvisualgeometrygrounded} & \textbf{13661*} & \textbf{223} & \textbf{1975} & \underline{21.74} & \underline{34.06} \\
        \midrule
        RayOcc (ours) & \underline{19251}* & \underline{243} & \underline{2620} & \textbf{22.03} & \textbf{34.84} \\
        \bottomrule 
    \addlinespace[0.4mm]
    \multicolumn{4}{l}{* denotes average number of Gaussians per scene.} \\
    \addlinespace[0.4mm]
    \multicolumn{5}{l}{The best and second-best performances denoted by \textbf{bold} and \underline{underline}.} \\
    \end{tabular}}
    \label{table:efficiency}
    \vspace{-3mm}
\end{table}

\subsection{Ablation Study}
We analyze RayOcc by comparing against representative Gaussian-based occupancy methods and by ablating the Gaussian sampling strategy.

\textbf{Sampling Methods Comparison.}
Table~\ref{table:initializing} evaluates the quality of Gaussian initialization using two position metrics. \emph{Perc.} measures the percentage of Gaussians placed inside occupied voxels, and \emph{Dist.} measures the average L1 distance to the nearest occupied voxel center.

GaussianFormer~\cite{huang2024gaussian} initializes Gaussian centers randomly in 3D space and relies on 4 iterative refinement to gradually move primitives toward occupied regions, which can waste representational capacity in early stages.
GaussianFormer-2~\cite{huang2024probabilistic} predicts a per-ray categorical depth distribution by training ray logits including an explicit empty logit with occupancy supervision, and then applies a softmax over depth to obtain a normalized distribution.
During Gaussian initialization, it samples a single depth bin per ray from this categorical distribution, introducing mutual competition across depth bins and restricting each ray to a single hypothesis.
VG3T~\cite{kim2025vg3tvisualgeometrygrounded} performs depth regression to obtain a per-pixel depth estimate and initializes Gaussians at the regressed depths, which also yields one dominant hypothesis per ray.

Compared with categorical depth sampling in GaussianFormer-2~\cite{huang2024probabilistic}, RayOcc increases Perc. from 28.85\% to 61.07\% and reduces Dist. from 1.24\,m to 0.70\,m, indicating more accurate and compact initialization.
Compared with VG3T~\cite{kim2025vg3tvisualgeometrygrounded}, RayOcc improves Perc. by 9.85 percentage points and reduces Dist. by 0.27\,m. These results indicate that component-wise multi-hypothesis initialization places Gaussian primitives closer to occupied regions than prior Gaussian initialization strategies.

\textbf{Efficiency Comparison.}
Table~\ref{table:efficiency} reports the number of Gaussians, latency, memory usage, and occupancy accuracy. Compared with GaussianFormer-2~\cite{huang2024probabilistic}, RayOcc improves IoU from 30.56 to 34.84 and mIoU from 20.02 to 22.03, while reducing latency from 513\,ms to 243\,ms and memory usage from 3063\,MB to 2620\,MB. Compared with VG3T~\cite{kim2025vg3tvisualgeometrygrounded}, RayOcc improves IoU from 34.06 to 34.84 and mIoU from 21.74 to 22.03, while increasing the average number of Gaussians from 13,661 to 19,251.

This reflects the intended trade-off of the proposed multi-hypothesis initialization. RayOcc allocates additional primitives to multiple plausible depth hypotheses rather than committing each ray to a single dominant hypothesis. These additional primitives improve Gaussian initialization quality and overall occupancy accuracy, at the cost of moderate increases in Gaussian count, latency, and memory compared with VG3T.

\begin{table}
    \centering
    \caption{Ablation Study on RayOcc's Adaptive Sampling}
    \setlength{\tabcolsep}{0.01\linewidth}
    \resizebox{1\linewidth}{!}{
    \begin{tabular}{c|cc|c|cc}
        \toprule
        \multirow{2}{*}{Sampling} & \multirow{2}{*}{mIoU} & \multirow{2}{*}{IoU} & \multirow{2}{*}{\makecell{Average Number \\ of Gaussians} $\downarrow$} & \multicolumn{2}{c}{Position} \\
        & & & & Perc. (\%) \( \uparrow \) & Dist. (m) \( \downarrow \) \\
        \midrule
        Bernoulli & 21.92 & \textbf{35.08} & 23136 & 53.55 & 0.85  \\
        $\tau=0.5$ & 22.01 & 34.90 & 19822 & 59.83 & 0.72 \\
        $\tau=0.7$ & 22.02 & 34.87 & 19681 & 60.41 & 0.71 \\
        $\tau=0.8$ & \textbf{22.03} & 34.84 & 19251 & 61.07 & 0.70 \\
        $\tau=0.9$ & \textbf{22.03} & 34.76 & \textbf{18679} & \textbf{62.38} & \textbf{0.68} \\
        \bottomrule
    \end{tabular}}
    \label{table:sampling_ablation}
    \vspace{-3mm}
\end{table}

\begin{figure*}
  \centering
  \includegraphics[width=1.0\textwidth]{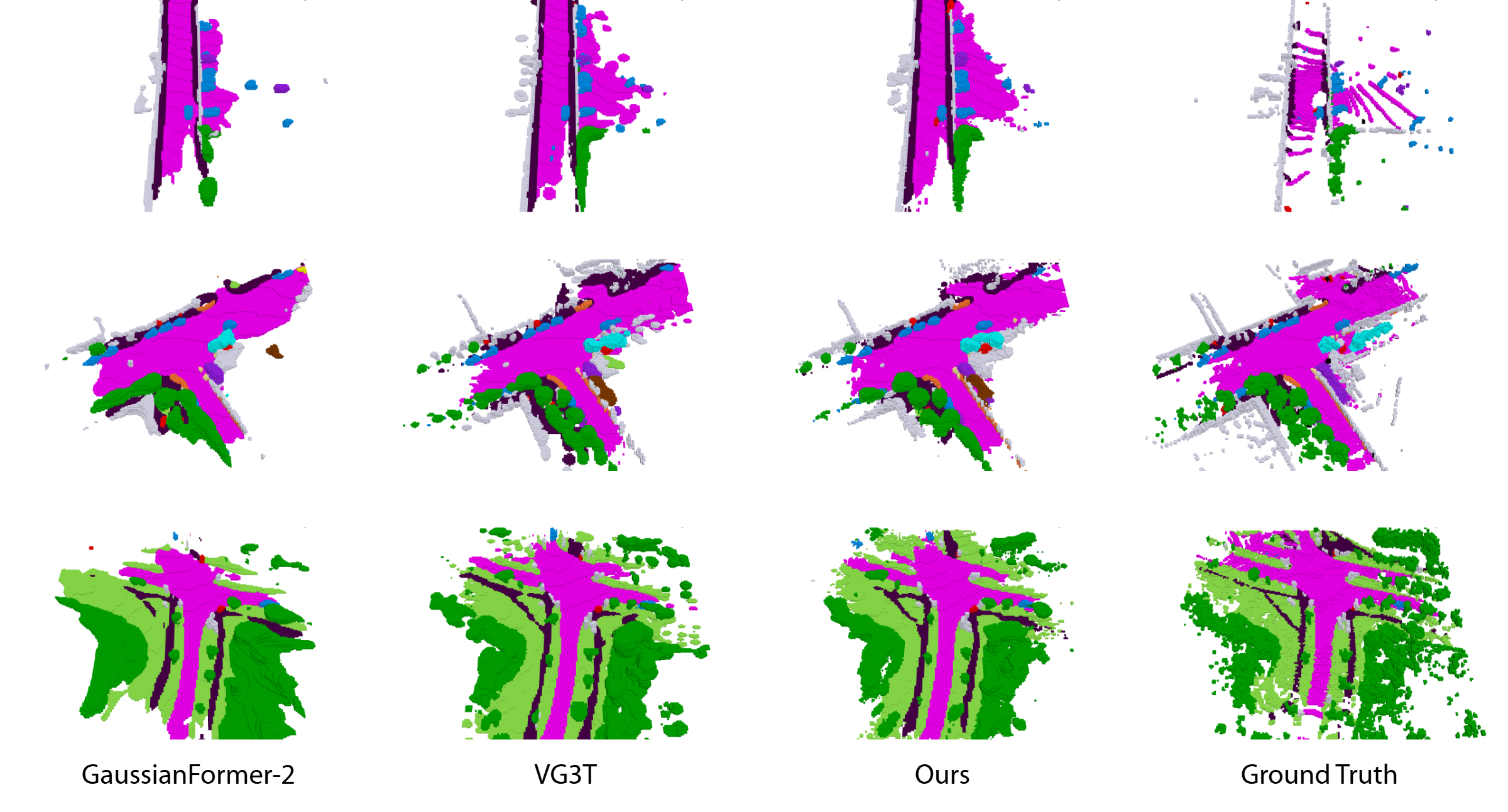}
  \caption{Qualitative comparison of predicted occupancy between our method with GaussianFormer-2 and VG3T.}
  \label{fig:fig4}
  \vspace{-3mm}
\end{figure*}

\textbf{Gaussian Sampling Strategy.}
We further study how the component activation rule affects both initialization quality and downstream occupancy prediction.
In Table~\ref{table:sampling_ablation}, $\tau$ denotes the deterministic activation threshold applied to component weights at inference, which directly controls the number of instantiated Gaussians per scene.
As $\tau$ increases, low-confidence components are progressively filtered out, leading to fewer Gaussians and consistently improved position quality.

Importantly, occupancy accuracy remains largely stable over a wide range of thresholds (0.7--0.9), indicating that RayOcc is not overly sensitive to the exact choice of $\tau$.
We use $\tau=0.8$ by default since it offers a favorable balance, it maintains the best mIoU while achieving higher IoU than $\tau=0.9$ with only a modest increase in Gaussian count.
When $\tau$ is set too low, the model instantiates more primitives, which can increase false positives and degrade position metrics. When $\tau$ is too high, aggressive pruning may under-cover thin structures or distant surfaces and slightly reduce IoU.

Finally, we report stochastic Bernoulli activation during training as a contrast to deterministic thresholding.
Stochastic activation encourages exploration of multiple hypotheses and provides gradients to the activation parameters, but it also introduces sampling variance and tends to generate more Gaussians.
For evaluation, we therefore adopt deterministic thresholding to obtain stable primitive counts and reproducible predictions.

\subsection{Visualization}
Figure~\ref{fig:fig3} visualizes the predicted 3D Gaussians and the resulting semantic occupancy for VG3T~\cite{kim2025vg3tvisualgeometrygrounded} and our RayOcc.
Overall, RayOcc produces cleaner and more spatially consistent occupancy, particularly in regions affected by occlusion.
For instance, in the first row RayOcc recovers the truck more coherently even when it is adjacent to fence-like structures, whereas VG3T tends to spill primitives into the fence region.
RayOcc also yields clearer ground layout and less fragmented occupancy around object boundaries, reducing spurious predictions near thin structures.
We attribute these improvements to our occlusion-aware ray occupancy formulation, which supports multiple depth hypotheses along a single line of sight and thus avoids over-committing to a single dominant depth estimate under severe ray-depth ambiguity.
Consequently, RayOcc allocates primitives more effectively to geometrically plausible surfaces, improving coverage in occluded scenes and producing higher-quality semantic occupancy outputs.
These qualitative observations are consistent with the quantitative gains in Table~\ref{table:nuscenes_results}.

Figure~\ref{fig:fig4} further provides a qualitative comparison of predicted occupancy among RayOcc, GaussianFormer-2~\cite{huang2024probabilistic}, and VG3T~\cite{kim2025vg3tvisualgeometrygrounded}.
Compared with GaussianFormer-2, RayOcc places more primitives in geometry occupied places, leading to less noisy occupancy and sharper reconstruction of planar structures such as building facades and road surfaces.
Compared with VG3T, RayOcc better preserves partially occluded objects and maintains more complete geometry behind occluders, reflecting the benefit of multi-label ray occupancy modeling.
Together, these visualizations highlight that RayOcc improves both the initialization of sparse Gaussians and the resulting dense occupancy prediction, especially in challenging scenes with heavy occlusion.

\section{CONCLUSIONS}

We presented RayOcc for addressing the mismatch between single-depth ray lifting and multi-label volumetric occupancy supervision in camera-only 3D semantic occupancy prediction. RayOcc models each ray with a non-normalized Gaussian mixture intensity and maps integrated intensity to interval-wise existence probabilities, allowing multiple occupancy hypotheses to coexist along a ray. The predicted mixture components are further converted into sparse 3D Gaussian primitives through adaptive component-wise sampling, improving Gaussian initialization under depth ambiguity. Experiments on nuScenes show improved overall occupancy accuracy and initialization quality.

\section*{ACKNOWLEDGMENT}
This work was supported by the National Research Foundation of Korea (NRF) grant funded by the Korea government (MSIT) (No. RS-2026-25497410). This work was also supported by the National Research Foundation of Korea (NRF) grant funded by the Korea government (MSIT) (No. RS-2026-25522067). This work was also supported by the Institute of Information \& Communications Technology Planning \& Evaluation (IITP) grant funded by the Korea government (MSIT) (No. RS-2025-02219317, AI Star Fellowship, Kookmin University).

\bibliographystyle{IEEEtran}
\bibliography{ref}

\end{document}